\documentclass{article}

\usepackage[preprint]{neurips_2024}

\usepackage[utf8]{inputenc} 
\usepackage[T1]{fontenc}    
\usepackage{hyperref}       
\usepackage{url}            
\usepackage{booktabs}       
\usepackage{amsfonts}       
\usepackage{nicefrac}       
\usepackage{microtype}      
\usepackage{xcolor}         
\usepackage{listings}%
\usepackage{adjustbox}%
\usepackage{graphicx}
\usepackage{amsmath}

\title{Aligning Motion-Blurred Images Using Contrastive Learning on Overcomplete Pixels}

\author{%
  Leonid Pogorelyuk \\
  Department of Mechanical, Aerospace, and Nuclear Engineering\\
  Rensselaer Polytechnic Institute\\
  Troy, NY 12180 \\
  \texttt{pogorl@rpi.edu} \\
  \And
  Stefan T.~Radev\\
  Center for Modeling, Simulation, \& Imaging in Medicine\\
  Rensselaer Polytechnic Institute\\
  Troy, NY 12180 \\
  \texttt{radevs@rpi.edu} \\
}

\newcommand*{\m}[1]{\mathbf{#1}}
\newcommand{\pb}{\boldsymbol{p}}

\begin{document}

\maketitle

\begin{abstract}
We propose a new contrastive objective for learning overcomplete pixel-level features that are invariant to motion blur. Other invariances (e.g., pose, illumination, or weather) can be learned by applying the corresponding transformations on unlabeled images during self-supervised training. We showcase that a simple U-Net trained with our objective can produce local features useful for aligning the frames of an unseen video captured with a moving camera under realistic and challenging conditions. Using a carefully designed toy example, we also show that the overcomplete pixels can encode the identity of objects in an image and the pixel coordinates relative to these objects.
\end{abstract}

\section{Introduction}

Establishing spatial correspondences between noisy images is a common problem in a wide variety of applications, from medicine \citep{chen2023image} to space navigation \citep{song2022deep}.
Such image alignment/matching problems are particularly challenging since they rarely come with known annotated ground truth correspondences, creating a strong impetus for developing and optimizing unsupervised methods.
Among these, \textit{self-supervised learning} (SSL) has gained significant attention due to its ability to pre-train models effectively for downstream tasks, often surpassing fully supervised methods in performance \citep{chen2020simple, he2020momentum, liu2021self}.
SSL effectively overcomes the need for ground-truth correspondences by creating auxiliary tasks, where the unlabeled data itself provides supervision to learn useful feature representations.

Most SSL methods focus on low-dimensional features or representations extracted from relatively high-dimensional images. The information distilled in these features is sufficient to tackle downstream tasks, such as generation or inference \citep{liu2021self}.
Ideally, the resulting features are invariant to various transformations, starting from hand-crafted features \citep{lowe1999object} invariant to scaling, and evolving into machine-learned features/keypoints invariant to more elaborate scenery changes, such as lighting or seasonal variations \citep{jing2022image}.

However, low-dimensional representations entail a significant loss of detail or granularity \citep{wang2021dense}.
This poses accuracy issues for image alignment and stitching problems, which require a much finer \textit{pixel-level} granularity. For this reason, recent SSL methods use low-dimensional features just for coarse alignment and rely on dedicated alignment networks for fine alignment and stitching \citep{sun2021loftr, lindenberger2023lightglue, xu2022gmflow, amir2021deep, jiang2021cotr,jia2023learning}.

In contrast, \textit{overcomplete} representations typically have latent dimensions that exceed the dimensions of the data, providing unique advantages for challenging vision tasks.
They increase robustness in the presence of noise \citep{yasarla2020exploring} and allow for much greater flexibility in matching structure in the data \citep{lewicki2000learning, wang2021dense}.
In the context of both human and machine learning \citep{dubova2023excess}, excess capacity representations have been found to support classification and predict learning speed \citep{tang2019effective}.
In dynamic systems, expansion of dimensionality can lead to more robust generalization and task adaptation in RNNs \citep{farrell2022gradient}.
Overcomplete representations have also been employed to learn unpaired shape transform of high-dimensional point clouds \citep{yin2019logan} and proven quintessential in hybrid models for challenging image processing problems, such as deep subspace clustering \citep{valanarasu2021overcomplete} and denoising of high-dimensional images \citep{yasarla2020exploring}.

In this work, we leverage overcomplete pixel embeddings to solve noisy alignment tasks.
We formulate a new self-supervised loss function
that does not require expensive student-teacher architectures and is potentially applicable to other downstream tasks, such as classification, image segmentation, and registration.
In summary, our contributions are:
\begin{enumerate}
    \item We formulate a new contrastive loss that induces overcomplete embeddings at the pixel level.
    \item We provide compelling image alignment results on a challenging task involving motion blur where standard feature-extraction methods fail.
    \item We construct a synthetic toy example that helps interpret the overcomplete pixels in the global context of an image.
\end{enumerate}

\section{Method}

\paragraph{Setup} 
The pose estimation problem consists of aligning two different camera views of the same world area taken under different conditions (such as time, illuminations, camera settings, as well as camera motion).
Each view is represented as an image indexed by a set of pixel coordinates $\pb \in \mathbb{N}^2$, such that $\m{I}(\pb) \in \mathbb{R}^C_I$ denotes the pixel value with $C_I$ channels indexed by $\pb$.
Our feature extractor $\m{F} = \mathcal{F}(\m{I})$ transforms an image $\m{I} \in \mathbb{R}^{H \times W \times C_I}$ into an overcomplete feature map $\m{F} \in \mathbb{R}^{H \times W \times C_F}$ with $C_F > C_I$.
Throughout the experiments, we use a standard fully-convolutional U-Net \citep{ronneberger2015u} architecture with $C_F$ output channels (see \textbf{Appendix A} for architecture details\footnote{Code is available at \url{https://github.com/leonidprinceton/oxels}.}), but transformer-based architectures, such as SegFormer \citep{xie2021segformer}, are also possible.

\begin{figure*}[t]
    \hspace{-0.3in}
    \includegraphics[trim=0 0.5in 0 0.1in, clip, width=6in]{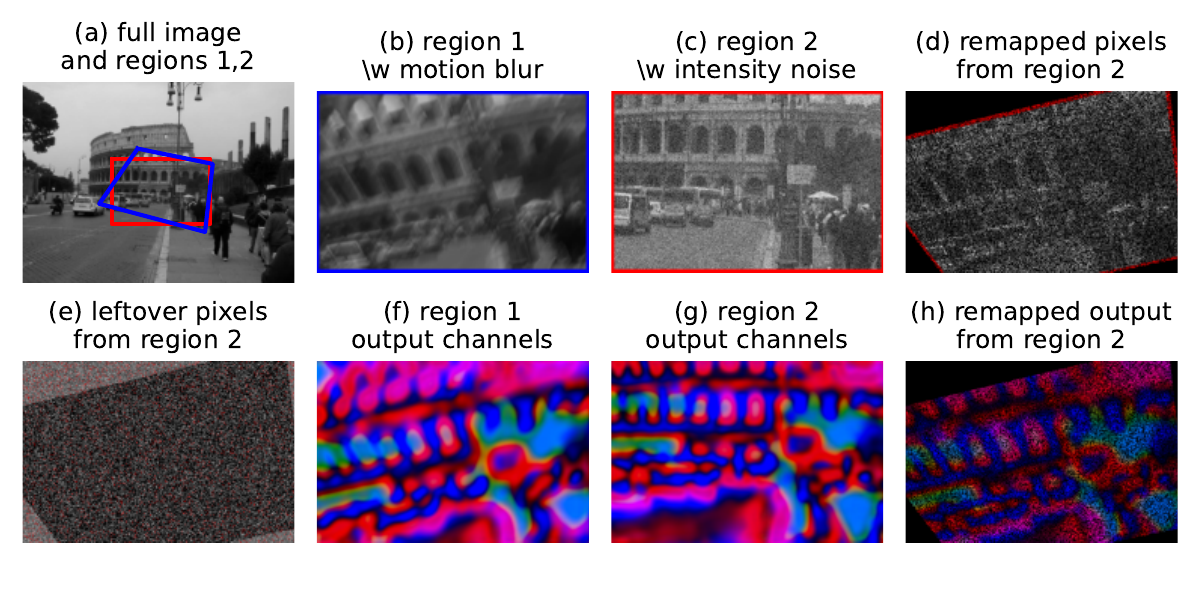} 
    \caption{\textbf{Training process illustration.} (a) Two regions of the same image (from the Image Matching Challenge 2022 \citep{image-matching-challenge-2022}) are selected. (b,c) After transforming the selected regions into rectangles, noise and motion blur are added to each. (d) A fraction of the pixel locations from region 2 is remapped (e.g., using \texttt{gather}) to align with the corresponding locations in region 1. (e) The rest of the pixel locations are randomly remapped. (f,g) Regions 1 and 2 are passed through the network to generate overcomplete features of the same image size but with more channels (just three channels are shown). (h) A fraction of the features of region 2 are remapped to align with features from region 1 and are encouraged to match in $L^{\infty}$. The rest of the features (not shown) are randomly remapped and encouraged not to match (contrastive loss).}
    \label{fig:tutorial}
\end{figure*}

\paragraph{Training}

In each training step, we sample two regions from image $k$ in the dataset, $\m{I}_{1,k}$ and $\m{I}_{2,k}$. Region $\m{I}_{2,k}$ is a random perspective transform of \smash{$\m{I}_{1,k}$} (see~\autoref{fig:tutorial}a) or a transform of the same camera view taken at a different time (including varying illumination conditions or moving objects). We also add various forms of artificial noise, such as color variations or motion blur (see~\autoref{fig:tutorial}b and c).

Next, we partition the pixels of each $\m{I}_{2,k}$ into a subset that maps pixels onto their corresponding positions in each $\m{I}_{1,k}$, and a subset that maps pixel onto a random location. 
We define $\mathcal{P}_k:\mathbb{N}^{2}\rightarrow\mathbb{N}^{2}$ as the approximate perspective map from pixels of $\m{I}_{2,k}$ to those of $\m{I}_{1,k}$, such that the two images would align (see~\autoref{fig:tutorial}d). 
Importantly, we do not align \textit{all} pixels, but instead select a fraction of all pixels using a function $\mathcal{S}_k: \mathbb{N}^2 \rightarrow \{-1, 1\}$. 
This function assigns a pixel the value of $\mathcal{S}_k(\pb) = 1$ if the pixel can be accurately mapped back from $\m{I}_{2,k}$ to $\m{I}_{1,k}$, and $\mathcal{S}_k(\pb) = -1$ otherwise (to be used in the contrastive part of the loss).
The pixels that cannot be accurately mapped are randomly mapped via \smash{$\mathcal{R}_k :\mathbb{N}^{2} \rightarrow \mathbb{N}^{2}$} (see~\autoref{fig:tutorial}e).
Thus, the process of partitioning the pixels of $\m{I}_{2,k}$ can be expressed as
\begin{equation}
\mathcal{M}_k(\pb) = \begin{cases}
\mathcal{P}_k(\pb) & \text{if}\quad \mathcal{S}_k(\pb) = 1\\
\mathcal{R}_k(\pb) & \text{if}\quad \mathcal{S}_k(\pb) = -1\\
\end{cases},
\end{equation}
where $\mathcal{M}_k$ maps all pixels between the two images.
The dependence on the index $k$ emphasizes that the transformation is different for each batch instance.
In practice, we train the feature extractor $\mathcal{F}$ on views of larger images (see~\autoref{fig:tutorial}a), but for brevity, we do not introduce new notation for these transformations.

\paragraph{Contrastive loss on overcomplete pixels}

The proposed loss works with the $L^{\infty}$ norm between the feature map of each original region $\m{I}_1$ at coordinates $\pb$ and that of $\m{I}_2$ at the transformed coordinates $\mathcal{M}_k(\pb)$:
\begin{equation}
\mathcal{D}_k(\pb) = \frac{1}{2}\left\Vert \mathcal{F}\left(\m{I}_{1,k}\right)(\pb) - \mathcal{F} \left(\m{I}_{2,k} \right)\left(\mathcal{M}_k(\pb)\right) \right\Vert _{\infty}.
\end{equation}
The per-pixel norms form the loss,
\begin{equation}
\mathcal{L} := \sum_{k}\sum_{\pb \in P}\left[ \mathcal{S}_k(\pb) \mathcal{D}_k(\pb) + \mathcal{D}_k(\pb)^2 \right]\label{eq:loss_func},
\end{equation}
where $P$ is a relevant set of pixels, which in our case is the set of all pixels $\pb = (i,j)$ for $i = 1,...,H$ and $j = 1,...,W$.
The term $\mathcal{S}_k(\pb)\mathcal{D}_k(\pb)$ encourages our overcomplete feature extractor $\mathcal{F}$ to decrease the distance between features that are expected to be the same ($S_k(\pb) = 1$), and increase the distance otherwise ($\mathcal{S}_k(\pb) = -1$). 
The second term, $\mathcal{D}_k(\pb)^2$, ensures that the loss does not grow unbounded. 
The $L^{\infty}$ norm encourages the feature channels to encode information in a discrete (bit-like) manner.

We control the ``weight'' of the contrastive part of the loss by assigning the probability of keeping the pixel $\pb$ aligned, $q = \mathrm{Pr}\left\{\mathcal{S}_k(\pb) = 1\right\}$. In our experiments, we found $q \sim 1/10$ to work well. Note that our approach does not require the standard contrastive loss part between different views across the batch $k'\ne k$ (see \textbf{Appendix B}).




\begin{figure*}[t]
    \hspace{-0.3in}
    \includegraphics[trim=0 0.1in 0 0.1in, clip, width=6in]{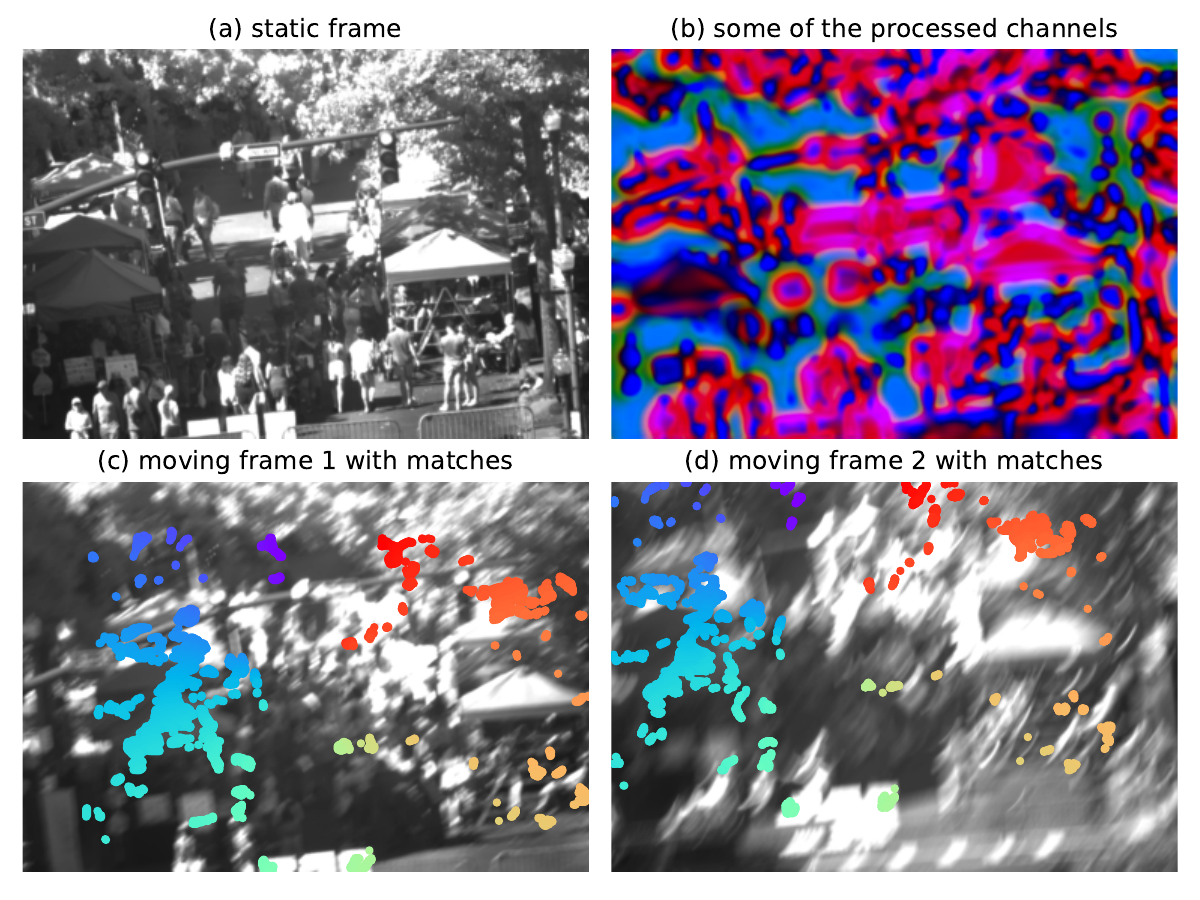} 
    \caption{\textbf{A challenging image alignment example.} (a) A static frame from a video of a farmers market (Troy, NY) taken from afar. (b) Three channels of the overcomplete representation generated by our network. (c and d) Frames 1 and 2 belong to a video captured with a moving global shutter camera. The frames are about 10 seconds apart and exhibit motion blur and overexposure. We applied a standard procedure of first finding the closest matches between the overcomplete features of the two frames, then using RANSAC to find matches corresponding to the same affine transformation between the two images. The colored dots correspond to all the locations successfully matched between the two frames despite the extreme motion blur.}
    \label{fig:application_example}
\end{figure*}

\begin{figure*}[t]
    \hspace{-0.3in}
    \includegraphics[trim=0 0.3in 0 0.1in, clip, width=6in]{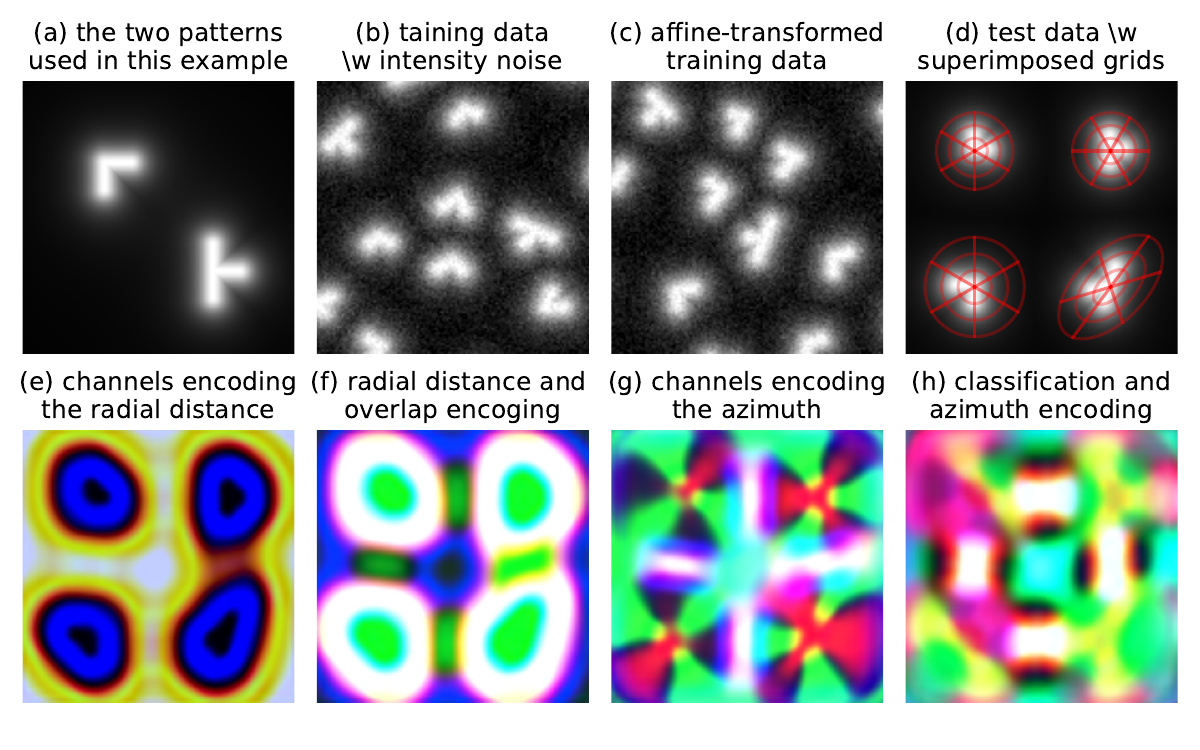} 
    \caption{\textbf{A toy example to interpret the overcomplete pixels learned by the network.} (a) Patterns A (top) and B (bottom) used in this example. (b and c) Two noisy training views of the same image composed of several randomly affine-transformed A or B patterns. (d) Two A and two B patterns with superimposed polar grids used to illustrate the affine transformation applied to the shapes. (e) Three out of twelve channels of the overcomplete output that encode information that resembles the radial distance from the centers of the shapes in pre-transformation coordinates. (f) More channels that encode radial distance and some of the overlap between adjacent areas (yellow areas). (g) Channels that encode mostly the polar angle (azimuth) by sectors similar to straight red lines in (d). (h) Channels that encode whether the center of the shape belongs to pattern A (red/magenta) or B (yellow/green), as well as some azimuth information.}
    \label{fig:toy_example}
\end{figure*}

\section{Empirical Evaluation}

\subsection{\label{sub:alignment}Alignment of Images with Motion Blur}

The loss function in~\autoref{eq:loss_func} is designed to make the network output invariant to the data transformations provided during training. In this work, we train the network to find features invariant to affine transformations, brightness variations, and motion blur. Motion blur is particularly challenging because it introduces a very distinct and \textit{a priori} unknown signature to all regions of the image (see~\autoref{fig:application_example}) that needs to be ignored by the feature extraction mechanism. 
Since motion blur is effectively a convolution, the network must implicitly learn to solve and regularize an ill-posed deconvolution problem \citep{zhang2022deep}.

Our training data consists of 5790 images from the Image Matching Challenge 2022 \citep{image-matching-challenge-2022} split into 16 folders by scene. We randomly selected one image from each folder for each training batch and overlapping regions within each image (similar to~\autoref{fig:tutorial}), transforming them into $256\times384$ rectangles and adding noise (brightness, contrast, pixel-wise intensity, motion blur with a simulated convolution kernel). Each training batch contained non-overlapping regions from the same image and from different images.

We test our network's performance on a video taken by a telescope, capturing different areas of a farmers market (see~\autoref{fig:application_example}a) as it slews around. \autoref{fig:application_example}c and d show two typical $552\times804\times1$ monochromatic frames from the video taken about 10 seconds apart with noticeable motion blur. The network was applied 9 times to subsets of each frame to cover it fully. The result is a $552\times804\times32$ output mapping each pixel to its overcomplete representation (\autoref{fig:application_example}b).

To align two images, we iterate over all $552 \times 804$ features in frame 1 and find the closest corresponding feature in frame 2 in $L^2$ norm. This is done efficiently by first inserting all features of frame 2 into a HNSW graph \citep{malkov2018efficient}. The affine transformation to align the images is found via the classical RANSAC algorithm \citep{fischler1981random}. All the features that were matched by the ``correct'' affine transformation are shown in~\autoref{fig:application_example} (not connected by lines as would typically be illustrated due to the very large number of matches). This example clearly illustrates the ease with which our network finds a large number of matching keypoints between the two different frames of never-seen-before scenery despite the prominent motion blur.

\subsection{\label{sub:toy_example}Toy Example: Understanding Overcomplete Representations}

In this experiment, we provide a synthetic toy example for which we can study and interpret the features induced by our loss functions.
We train the same U-Net \citep{ronneberger2015u} on $128\times128\times1$ images of multiple objects that belong to one of two patterns A or B (shown in \autoref{fig:toy_example}a). The objects are randomly affine-transformed and scattered in each image, and the training data consists of two affine transforms (views) of the same image (see \autoref{fig:toy_example}b and c) with intensity noise. The network learns overcomplete representations with 12 channels (output shape $128\times128\times12$) via the loss function in ~\autoref{eq:loss_func}.

To interpret the learned features, we consider two pairs of A and B patterns with some affine transformation applied to them but no noise. \autoref{fig:toy_example}d shows the four shapes with polar grids superimposed on top of them for illustration purposes. The grids underwent the same affine transformations as their corresponding shapes. Hence, the grids represent the radial distance and polar angle relative to the original shapes (before they were scaled, rotated, etc.).

It appears that the network learns to encode the polar coordinates relative to each shape in the image and classify the shapes. \autoref{fig:toy_example}e shows three output channels, which form roughly ellipses centered and aligned with the shapes. \autoref{fig:toy_example}f shows channels that also encode the radial distance and potentially the overlap between shapes (shown in yellow). \autoref{fig:toy_example}g shows channels that encode the polar angle by splitting the area around each shape into sectors centered at the shape itself. \autoref{fig:toy_example}h shows channels that also classify the patterns (red+magenta colors at the center of the shape vs. green+yellow). We note that the distinction between polar coordinates and classification is not clear-cut. The polar encoding in \autoref{fig:toy_example}g depends on the shape involved, and there is some color overlap away from the shapes' center in \autoref{fig:toy_example}h.



\section{Conclusion}

We present a pixel-based cost function that encourages a network to learn overcomplete representations, including object/feature classification and relative positioning within images. That said, the overcomplete features may not be generally interpretable as in our toy example (\autoref{sub:toy_example}) since channels might get entangled. Even in our toy example, the channels encoding the polar angle do so in a shape-dependent manner (see, for example, the number of red sectors in \autoref{fig:toy_example}g).

The data diversity introduced during training encourages the network to learn features invariant to the different conditions present in the data. Applying perspective transformations, for example, makes the overcomplete features locally invariant to scaling and rotation.
Our approach can trivially be extended to other common invariances such as illumination and weather conditions up to the availability of properly aligned data.

In~\autoref{sub:alignment}, we focused on learning invariance to motion blur, which we artificially introduced when training on Image Matching Challenge 2022 data \citep{image-matching-challenge-2022}. We then used our network's overcomplete pixels to align the frames of a video with a moving camera feed of a street view that is somewhat dissimilar to the training dataset and has significant motion blur present.

\bibliography{references.bib}

\appendix

\section{Implementation Details and Hyperparameters}

All networks and numerical experiments are implemented in TensorFlow \citep{abadi2016tensorflow}.
Throughout, we use an Adam optimizer with an initial learning rate of 0.0005, default hyperparameters, global norm clipping between $[0, 1]$, and a cosine learning rate decay schedule that brings down the learning rate to zero after the final iteration (i.e., $\alpha = 0$).
All networks are trained on a single machine equipped with an NVIDIA\textsuperscript{\textregistered} RTX 4090 graphics accelerator with 24GB of GPU memory. 

For the alignment task in~\autoref{sub:alignment} we train a standard fully convolutional U-Net \citep{ronneberger2015u} of depth five with input size $256\times384\times1$ and output size $256\times384\times32$. The number of channels in each layer before the bottleneck increases from 8 to 256, and remains at 256 after the bottleneck (for a total of 29 million parameters). The final convolutional layer decreases the number of channels to $C_F = 32$.
We train the network for a total of 3000 epochs with 30 update steps per epoch and a batch size of 8.

The dataset in~\autoref{sub:alignment} consists of 5790 images from the Image Matching Challenge 2022 \citep{image-matching-challenge-2022}. If necessary, the images were re-scaled to a minimum height of 512 and width of 768.

For the toy example in~\autoref{sub:toy_example}, we use the same architecture with $128\times128\times1$ input, $128\times128\times12$ output, and channels increasing from 8 to 256.
We train the network for a total of 200 epochs with 1024 update steps per epoch and a batch size of 8.

\section{Negative Pairs}

We can augment our original loss with a contrastive component over ``negative examples'' that compares the dense features of different view pairs $k$ and $k'$ within a batch. 
The order of the pixels does not matter since any two feature vectors should differ in $L^{\infty}$,
\begin{equation}
\mathcal{C}_{k, k'}(\pb) = \left\Vert \mathcal{F} \left(\m{I}_{2,k} \right)(\pb) - \mathcal{F} \left(\m{I}_{2,k'} \right)(\pb) \right\Vert _{\infty}.
\end{equation}
The negative loss component is taken over all pairs of unrelated views $k\ne k'$ pertaining to the transformed images $\{\m{I}_{2,k}\}_{k=1}^K$:
\begin{equation}
\mathcal{L}_{\text{negative}} := \frac{1}{K^2}\sum_{k=1}^K\sum_{k'=1}^{k-1} \sum_{\pb \in P}\left[-\mathcal{C}_{k, k'}(\pb) + \mathcal{C}_{k, k'}(\pb)^2 \right].
\end{equation}
In this case, the total loss becomes,
\begin{equation}
   \mathcal{L}_{\text{total}} = \mathcal{L} + \lambda \cdot \mathcal{L}_{\text{negative}},
\end{equation}

where $\lambda$ is a contrastive weight hyperparameter. Our ablation studies showed no advantage of $\mathcal{L}_{\text{negative}}$ for the challenging image alignment task, but we discuss it for completeness.

\end{document}